\documentclass[runningheads]{llncs}
\usepackage{graphicx}
\begin{document}
\title{Attention-based method for categorizing different types of online harassment language}
\titlerunning{Attention-based method for categorizing online harassment language}
%
\author{Christos Karatsalos\and
Yannis Panagiotakis}

\institute{Athens University of Economics and Business, Athens, Greece \\
\email{\{ckarats, giannispanagiwtakis\}@gmail.com}}
\maketitle              
\begin{abstract}
In the era of social media and networking platforms, Twitter has been doomed for abuse and harassment toward users specifically women. Monitoring the contents including sexism and sexual harassment in traditional media is easier than monitoring on the online social media platforms like Twitter, because of the large amount of user generated content in these media. So, the research about the automated detection of content containing sexual or racist harassment is an important issue and could be the basis for removing that content or flagging it for human evaluation. Previous studies have been focused on collecting data about sexism and racism in very broad terms. However, there is no much study focusing on different types of online harassment attracting natural language processing techniques. In this work, we present an multi-attention based approach for the detection of different types of harassment in tweets. Our approach is based on the Recurrent Neural Networks and particularly we are using a deep, classification specific multi-attention mechanism. Moreover, we tackle the problem of imbalanced data, using a back-translation method. Finally, we present a comparison between different approaches based on the Recurrent Neural Networks.

\keywords{Text classification  \and Twitter \and Hate Speech \and Deep Learning \and Attention Mechanism.}
\end{abstract}
\section{Introduction}
In the era of social media and networking platforms, Twitter has been doomed for abuse and harassment toward users specifically women. In fact, online harassment becomes very common in Twitter and there have been a lot of critics that Twitter has become the platform for many racists, misogynists and hate groups which can express themselves openly. Online harassment is usually in the form of verbal or graphical formats and is considered harassment, because it is neither invited nor has the consent of the receipt. Monitoring the contents including sexism and sexual harassment in traditional media is easier than monitoring on the online social media platforms like Twitter. The main reason is because of the large amount of user generated content in these media. So, the research about the automated detection of content containing sexual harassment is an important issue and could be the basis for removing that content or flagging it for human evaluation. The basic goal of this automatic classification is that it  will significantly improve the process of detecting these types of hate speech on social media by reducing the time and effort required by human beings.

Previous studies have been focused on collecting data about sexism and racism in very broad terms or have proposed two categories of sexism as benevolent or hostile sexism~\cite{ref_article1}, which undermines other types of online harassment. However, there is no much study focusing on different types online harassment alone attracting natural language processing techniques. 

In this paper we present our work, which is a part of the SociaL Media And Harassment Competition of the ECML PKDD 2019 Conference. The topic of the competition is the classification of different types of harassment and it is divided in two tasks. The first one is the classification of the tweets in harassment and non-harassment categories, while the second one is the classification in specific harassment categories like indirect harassment, physical and sexual harassment as well.
We are using the dataset of the competition, which includes text from tweets having the aforementioned categories. Our approach is based on the Recurrent Neural Networks and particularly we are using a deep, classification specific attention mechanism. Moreover, we present a comparison between different variations of this attention-based approach like multi-attention and single attention models. The next Section includes a short description of the related work, while the third Section includes a description of the dataset. After that, we describe our methodology. Finally, we describe the experiments and we present the results and our conclusion.

\section{Related Work}
Waseem et al.~\cite{ref_article2} were the first who collected hateful tweets and categorized them into being sexist, racist or neither. However, they did not provide specific definitions for each category. Jha and Mamidi ~\cite{ref_article1} focused on just sexist tweets and proposed two categories of hostile and benevolent sexism. However, these categories were general as they ignored other types of sexism happening in social media. Sharifirad S. and Matwin S.~\cite{ref_article3}  proposed complimentary categories of sexist language inspired from social science work. They categorized the sexist tweets into the categories of indirect harassment, information threat, sexual harassment and physical harassment. In the next year the same authors proposed~\cite{ref_article4} a more comprehensive categorization of online harassment in social media e.g. twitter into the following categories, indirect harassment, information threat, sexual harassment, physical harassment and not sexist.

For the detection of hate speech in social media like twitter, many approaches have been proposed. Jha and Mamidi~\cite{ref_article1} tested support vector machine, bi-directional RNN encoder-decoder and FastText on hostile and benevolent sexist tweets. They also used SentiWordNet and subjectivity lexicon on the extracted phrases to show the polarity of the tweets. Sharifirad et al.~\cite{ref_article5} trained, tested and evaluated different classification methods on the SemEval2018 dataset and chose the classifier with the highest accuracy for testing on each category of sexist tweets to know the mental state and the affectual state of the user who tweets in each category. To overcome the limitations of small data sets on sexist speech detection, Sharifirad S. et al.~\cite{ref_article6} have applied text augmentation and text generation with certain success. They have generated new tweets by replacing words in order to increase the size of our training set. Moreover, in the presented text augmentation approach, the number of tweets in each class remains the same, but their words are augmented with words extracted from their ConceptNet relations and their description extracted from Wikidata. Zhang et al.~\cite{ref_article7} combined convolutional and gated recurrent networks to detect hate speech in tweets. Others have proposed different methods, which are not based on deep learning. Burnap and Williams~\cite{ref_article8} used Support Vector Machines, Random Forests and a meta-classifier to distinguish between hateful and non-hateful messages. A survey of recent research in the field is presented in~\cite{ref_article9}. For the problem of the hate speech detection a few approaches have been proposed that are based on the Attention mechanism. Pavlopoulos et al.~\cite{ref_article10} have proposed a novel, classification-specific attention mechanism that improves the performance of the RNN further for the detection of abusive content in the web. Xie et al.~\cite{ref_article11} for emotion intensity prediction, which is a similar problem to ours, have proposed a novel attention mechanism for CNN model that associates attention-based weights for every convolution window. Park and Fung~\cite{ref_article14} transformed the classification into a 2-step problem, where abusive text ﬁrst is distinguished from the non-abusive, and then the class of abuse (Sexism or Racism) is determined. However, while the first part of the two step classification performs quite well, it falls short in detecting the particular class the abusive text belongs to. Pitsilis et al.~\cite{ref_article15} have proposed a detection scheme that is an ensemble of RNN classiﬁers, which incorporates various features associated with user related information, such as the users’ tendency towards racism or sexism
\begin{table}[t]
\caption{Class distribution of the dataset.}\label{tab1}
\begin{tabular}{|l|l|l|l|l|l|l|}
\hline
Dataset    & Tweets & Harassment & Harassment(\%) & Indirect (\%) & Sexual (\%) & Physical (\%) \\
\hline
train      & 6374   & 2713       & 42.56         & 0.86        & 40.50      & 1.19         \\
\hline
validation & 2125   & 632        & 29.74          & 3.34         & 24.76     & 1.69         \\
\hline
test       & 2123   & 611        & 28.78         & 9.28         & 14.69     & 4.71        \\
\hline
\end{tabular}
\end{table}

\section{Dataset description}
The dataset from Twitter that we are using in our work, consists of a train set, a validation set and a test set. It was published for the "First workshop on categorizing different types of online harassment languages in social media". The whole dataset is divided into two categories, which are harassment and non-harassment tweets. Moreover, considering the type of the harassment, the tweets are divided into three sub-categories which are indirect harassment, sexual and physical harassment. We can see in Table~\ref{tab1} the class distribution of our dataset. One important issue here is that the categories of indirect and physical harassment seem to be more rare in the train set than in the validation and test sets. To tackle this issue, as we describe in the next section, we are performing data augmentation techniques. However, the dataset is imbalanced and this has a significant impact in our results.

\section{Proposed methodology}
\begin{figure}[t]
    \centering
    \includegraphics[width=6cm]{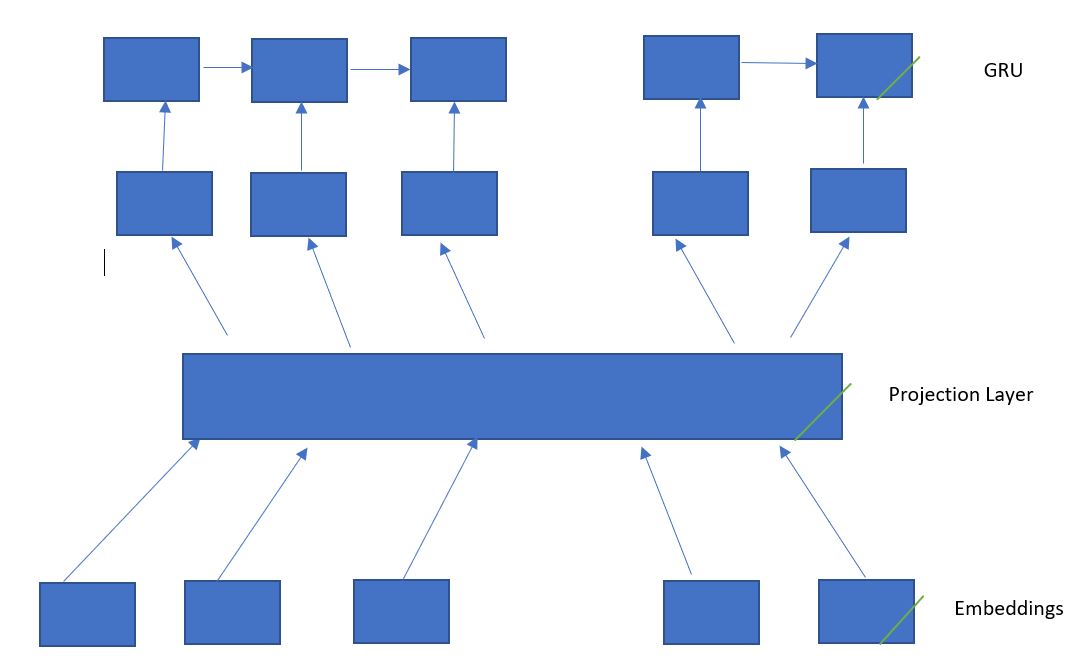}
    \caption{Projection Layer}
    \label{fig:projectionLayer}
\end{figure}

\subsection{Data augmentation}
As described before one crucial issue that we are trying to tackle in this work is that the given dataset is imbalanced. Particularly, there are only a few instances from indirect and physical harassment categories respectively in the train set, while there are much more in the validation and test sets for these categories. To tackle this issue we applying a back-translation method~\cite{ref_article16}, where we translate indirect and physical harassment tweets of the train set from english to german, french and greek. After that, we translate them back to english in order to achieve data augmentation. These "noisy" data that have been translated back, increase the number of indirect and physical harassment tweets and boost significantly the performance of our models.

Another way to enrich our models is the use of pre-trained word embeddings from 2B Twitter data~\cite{ref_article17} having 27B tokens, for the initialization of the embedding layer.
\subsection{Text processing}
Before training our models we are processing the given tweets using a tweet pre-processor\footnote[1]{\url{https://pypi.org/project/tweet-preprocessor/}}. The scope here is the cleaning and tokenization of the dataset. 
\subsection{RNN Model and Attention Mechanism}
\begin{figure}[t]
    \centering
    \includegraphics[width=4cm]{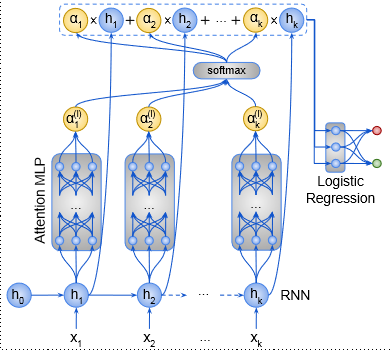}
    \caption{Attention mechanism, MLP with l Layers}
    \label{fig:attentionMechanism}
\end{figure}
We are presenting an attention-based approach for the problem of the harassment detection in tweets. In this section, we describe the basic approach of our work. We are using RNN models because of their ability to deal with sequence information. The RNN model is a chain of GRU cells~\cite{ref_article18} that transforms the tokens \(w_{1}, w_{2},..., w_{k}\) of each tweet to the hidden states \(h_{1}, h_{2},..., h_{k}\), followed by an LR Layer that uses \(h_{k}\) to classify the tweet as harassment or non-harassment (similarly for the other categories). Given the vocabulary V and a matrix E $\in$ \(R^{d \times \vert V \vert}\) containing d-dimensional word embeddings, an initial \(h_{0}\) and a tweet $w = <w_{1},.., w_{k}>$, the RNN computes \(h_{1}, h_{2},..., h_{k}\), with $h_{t} \in R^{m}$, as follows:  
\[h^{'}_{t} = \tanh(W_{h}x_{t} + U_{h}(r_{t}\odot h_{t-1}) + b_{h})\]
\[h_{t} = (1-z_{t})\odot h_{t-1} + z_{t}\odot h^{'}_{t}\]
\[z_{t} = \sigma(W_{z}x_{t} + U_{z}h_{t-1} + b_{z})\]
\[r_{t} = \sigma(W_{r}x_{t} + U_{r}h_{t-1} + b_{r})\]
where $h^{'}_{t} \in R^{m}$ is the proposed hidden state at position t, obtained using the word embedding $x_{t}$ of token $w_{t}$ and the previous hidden state $h_{t-1}$, $\odot$ represents the element-wise multiplication, $r_{t} \in R^{m}$ is the reset gate, $z_{t} \in R^{m}$ is the update gate, $\sigma$ is the sigmoid function. Also $W_{h}, W_{z}, W_{r} \in R^{m \times d}$ and $U_{h}, U_{z}, U_{r} \in R^{m \times m}$, $b_{h}, b_{z}, b_{r} \in R^{m}$. After the computation of state $h_{k}$ the LR Layer estimates the probability that tweet w should be considered as harassment, with $W_{p} \in R^{1 \times m}, b_{p} \in R$: 
\[P_{RNN}(harassment \vert w) = \sigma (W_{p}h_{k} + b_{p}).\]
We would like to add an attention mechanism similar to the one presented in~\cite{ref_article10}, so that the LR Layer will consider the weighted sum $h_{sum}$ of all the hidden states instead of $h_{k}$:
\begin{center}
$h_{sum} = \sum_{t=1}^{k} \alpha_{t}h_{t}$

$P_{attentionRNN} = \sigma (W_{p}h_{sum} + b_{p})$
\end{center}
Alternatively, we could pass $h_{sum}$ through an MLP with k layers and then the LR layer will estimate the corresponding probability. More formally, 
\begin{center}
$P_{attentionRNN} = \sigma (W_{p}h_{*} + b_{p})$
\end{center}
where $h_{*}$ is the state that comes out from the MLP. The weights $\alpha_{t}$ are produced by an attention mechanism presented in~\cite{ref_article10} (see Fig.~\ref{fig:attentionMechanism}), which is an MLP with l layers. This attention mechanism differs from most previous ones~\cite{ref_article19,ref_article20}, because it is used in a classification setting, where there is no previously generated output sub-sequence to drive the attention. It assigns larger weights $\alpha_{t}$ to hidden states $h_{t}$ corresponding to positions, where there is more evidence that the tweet should be harassment (or any other specific type of harassment) or not. In our work we are using four attention mechanisms instead of one that is presented in~\cite{ref_article10}. Particularly, we are using one attention mechanism per category. Another element that differentiates our approach from Pavlopoulos et al.~\cite{ref_article10} is that we are using a projection layer for the word embeddings (see Fig.~\ref{fig:projectionLayer}). In the next subsection we describe the Model Architecture of our approach.
\subsection{Model Architecture}
The Embedding Layer is initialized using pre-trained word embeddings of dimension 200 from Twitter data that have been described in a previous sub-section. After the Embedding Layer, we are applying a Spatial Dropout Layer, which drops a certain percentage of dimensions from each word vector in the training sample. The role of Dropout is to improve generalization performance by preventing activations from becoming strongly correlated~\cite{ref_article13}. Spatial Dropout, which has been proposed in~\cite{ref_article12}, is an alternative way to use dropout with convolutional neural networks as it is able to dropout entire feature maps from the convolutional layer which are then not used during pooling. After that, the word embeddings are passing through a one-layer MLP, which has tanh as activation function and 128 hidden units, in order to project them in the vector space of our problem considering that they have been pre-trained using text that has a different subject. In the next step the embeddings are fed in a unidirectional GRU having 1 Stacked Layer and size 128. We prefer GRU than LSTM, because it is more efficient computationally. Also the basic advantage of LSTM which is the ability to keep in memory large text documents, does not hold here, because tweets supposed to be not too large text documents. The output states of the GRU are passing through four self-attentions like the one described above~\cite{ref_article10}, because we are using one attention per category (see Fig.~\ref{fig:attentionMechanism}). Finally, a one-layer MLP having 128 nodes and ReLU as activation function computes the final score for each category. At this final stage we have avoided using a softmax function to decide the harassment type considering that the tweet is a harassment, otherwise we had to train our models taking into account only the harassment tweets and this might have been a problem as the dataset is not large enough.

\section{Experiments}
\subsection{Training Models}
In this subsection we are giving the details of the training process of our models. Moreover, we are describing the different models that we compare in our experiments. 

Batch size which pertains to the amount of training samples to consider at a time for updating our network weights, is set to 32, because our dataset is not large and small batches might help to generalize better. Also, we set other hyperparameters as: epochs = 20, patience = 10. As early stopping criterion we choose the average AUC, because our dataset is imbalanced.

The training process is based on the optimization of the loss function mentioned below and it is carried out with the Adam optimizer~\cite{ref_article21}, which is known for yielding quicker convergence. We set the learning rate equal to 0.001: 
\begin{center}
$L = \frac{1}{2}BCE(harassment) + \frac{1}{2}(\frac{1}{5}BCE(sexualH) + \frac{2}{5}BCE(indirectH)+\frac{2}{5}BCE(physicalH))$
\end{center}
where BCE is the binary cross-entropy loss function,
\begin{center}
$BCE = -\frac{1}{n}\sum_{i=1}^{n}[y_{i}log(y^{'}_{i}) + (1 - y_{i})log(1 - y^{'}_{i}))]$
\end{center}
$i$ denotes the $i$th training sample, $y$ is the binary representation of true harassment label, and $y^{'}$ is the predicted probability.
In the loss function we have applied equal weight to both tasks. However, in the second task (type of harassment classification) we have applied higher weight in the categories that it is harder to predict due to the problem of the class imbalance between the training, validation and test sets respectively.
\subsection{Evaluation and Results}
\begin{table}[t]
\caption{The results considering F1 Score.}\label{tab2}
\resizebox{\textwidth}{!} {
\begin{tabular}{|l|l|l|l|l|l|}
\hline
\textbf{Model}    & \textbf{sexual\_f1} & \textbf{indirect\_f1} & \textbf{physical\_f1} & \textbf{harassment\_f1} & \textbf{f1\_macro} \\
\hline
attentionRNN     & 0.674975   & 0.296320       & 0.087764         & 0.709539        & 0.442150         \\
\hline
MultiAttentionRNN & 0.693460   & 0.325338        & 0.145369          & 0.700354         & 0.466130         \\
\hline
MultiProjectedAttentionRNN      & 0.714094   & 0.355600        & 0.126848         & 0.686694         & \textbf{0.470809}        \\
\hline
ProjectedAttentionRNN      & 0.692316   & 0.315336        & 0.019372         & 0.694082         & 0.430276        \\
\hline
AvgRNN      & 0.637822   & 0.175182        & 0.125596         & 0.688122         & 0.40668        \\
\hline
LastStateRNN      & 0.699117   & 0.258402        & 0.117258         & 0.710071         & 0.446212        \\
\hline
ProjectedAvgRNN      & 0.655676   & 0.270162        & 0.155946         & 0.675745         & 0.439382        \\
\hline
ProjectedLastStateRNN      & 0.696184   & 0.334655        & 0.072691         & 0.707994         & 0.452881        \\
\hline
\end{tabular}
}
\end{table}
Each model produces four scores and each score is the probability that a tweet includes harassment language, indirect, physical and sexual harassment language respectively. For any tweet, we first check the score of the harassment language and if it is less than a specified threshold, then the harassment label is zero, so the other three labels are zero as well. If it is greater than or equal to that threshold, then the harassment label is one and the type of harassment is the one among these three having that has the greatest score (highest probability). We set this threshold equal to 0.33. 

We compare eight different models in our experiments. Four of them have a Projected Layer (see Fig.~\ref{fig:projectionLayer}), while the others do not have, and this is the only difference between these two groups of our models. So, we actually include four models in our experiments (having a projected layer or not). Firstly, LastStateRNN is the classic RNN model, where the last state passes through an MLP and then the LR Layer estimates the corresponding probability. In contrast, in the AvgRNN model we consider the average vector of all states that come out of the cells. The AttentionRNN model is the one that it has been presented in~\cite{ref_article10}. Moreover, we introduce the MultiAttentionRNN model for the harassment language detection, which instead of one attention, it includes four attentions, one for each category. 

We have evaluated our models considering the F1 Score, which is the harmonic mean of precision and recall. We have run ten times the experiment for each model and considered the average F1 Score. The results are mentioned in Table~\ref{tab2}. Considering F1 Macro the models that include the multi-attention mechanism outperform the others and particularly the one with the Projected Layer has the highest performance. In three out of four pairs of models, the ones with the Projected Layer achieved better performance, so in most cases the addition of the Projected Layer had a significant enhancement.

\section{Conclusion - Future work}
We present an attention-based approach for the detection of harassment language in tweets and the detection of different types of harassment as well. Our approach is based on the Recurrent Neural Networks and particularly we are using a deep, classification specific attention mechanism. Moreover, we present a comparison between different variations of this attention-based approach and a few baseline methods. According to the results of our experiments and considering the F1 Score, the multi-attention method having a projected layer, achieved the highest performance. Also, we tackled the problem of the imbalance between the training, validation and test sets performing the technique of back-translation.

In the future, we would like to perform more experiments with this dataset applying different models using BERT~\cite{ref_article22}. Also, we would like to apply the models presented in this work, in other datasets about hate speech in social media.

%
%
%

\begin{thebibliography}{8}
\bibitem{ref_article1}
Jha, A., Mamidi, R. (2017). When does a compliment become sexist: Analysis and classification of ambivalent sexism using twitter data. Proceedings of the Second Workshop on Natural Language Processing and Computational Social Science

\bibitem{ref_article2}
Waseem, Z., Hovy, D. 2016. Hateful symbols or hateful people: Predictive features for hate speech detection on twitter. In Proceedings of NAACL-HLT. 88–93.  

\bibitem{ref_article3}
Sharifirad S. and Matwin S, 2018.classification of Different Types of Sexist Languages on Twitter and the Gender Footprint on Each of the Classes, CICLing 2018.

\bibitem{ref_article4}
Sharifirad S. and Matwin S. 2019. When a Tweet is Actually Sexist. A more Comprehensive classification of Different Online Harassment Categories and The Challenges in NLP:  \url{https://arxiv.org/abs/1902.10584}

\bibitem{ref_article5}
Sharifirad S., Matwin S. and Jafarpour B. 2019. How is Your Mood When Writing Sexist tweets? Detecting the Emotion Type and Intensity of Emotion Using Natural Language Processing Techniques: \url{https: //arxiv.org/abs/1902.03089}

\bibitem{ref_article6}
Sharifirad S., Matwin S. and Jafarpour B. 2018. Boosting Text classification Performance on Sexist Tweets by Text Augmentation and Text Generation Using a Combination of Knowledge Graphs: \url{http://aclweb.org/anthology/W18-5114}

\bibitem{ref_article7}
Ziqi Zhang, David Robinson, and Jonathan Tepper. 2018. Detecting Hate Speech on Twitter Using a ConvolutionGRU Based Deep Neural Network. In Lecture Notes in Computer Science. ESWC 2018, Heraklion, Greece.

\bibitem{ref_article8}
Burnap P. and Williams M. 2015. Cyber hate speech on twitter: An application of machine classification and statistical modeling for policy and decision making. Policy and Internet, 7(2):223–242.

\bibitem{ref_article9}
Schmidt A. and Wiegand M. 2017. A Survey on Hate Speech Detection Using Natural Language Processing. In Proceedings of the Fifth International Workshop on Natural Language Processing for Social Media. Association for Computational Linguistics, pages 1–10, Valencia, Spain.

\bibitem{ref_article10}
Pavlopoulos, J., Malakasiotis, P. and Androutsopoulos, I. (2017). Deeper attention to abusive user content moderation. In Proceedings of the 2017 Conference on Empirical Methods in Natural Language Processing, pp. 1125–1135. Association for Computational Linguistics.

\bibitem{ref_article11}
Xie H., Feng S., Wang D., Zhang Y. (2018). A Novel Attention Based CNN Model for Emotion Intensity Prediction. In: Zhang M., Ng V., Zhao D., Li S., Zan H. (eds) Natural Language Processing and Chinese Computing. NLPCC 2018. Lecture Notes in Computer Science, vol 11108. Springer, Cham.

\bibitem{ref_article12}
Tompson J., Goroshin R., Jain A., LeCun Y., and Bregler C. (2015). Efficient object localization using convolutional networks. In The IEEE Conference on Computer Vision and
Pattern Recognition (CVPR), June 2015.

\bibitem{ref_article13}
Hinton G.E., Srivastava N., Krizhevsky A., Sutskever I.,and  Salakhutdinov R. R. (2012). Improving neural networks by preventing co-adaptation of feature detectors. arXiv preprint arXiv:1207.0580, 2012.

\bibitem{ref_article14}
Park J.H. and Fung P. 2017. One-step and two-step classification for abusive language detection on twitter. 1st Workshop on Abusive Language Online, ACL 2017, VancouverCanada, 4th August 2017.

\bibitem{ref_article15}
Pitsilis G., Ramampiaro H., and Langseth H. 2018. Detecting offensive language in tweets using deep learning. arXiv preprint arXiv:1801.04433.

\bibitem{ref_article16}
Sennrich R., Haddow B., and Birch A. 2016. Improving neural machine translation models with monolingual data. In Proceedings of the 54th Annual Meeting of the Association for Computational Linguistics (Volume 1: Long Papers), pp. 86–96, Berlin, Germany, August 2016a. Association for Computational Linguistics.

\bibitem{ref_article17}
Pennington J., Socher R., Manning C.D. 2014. “GloVe: Global Vectors for Word Representation”, Proceedings of the 2014 Conference on Empirical Methods in Natural Language Processing (EMNLP), Doha, Qatar, 1532–1543.

\bibitem{ref_article18}
Cho K., van Merrienboer B., Gulcehre C., Bahdanau D., Bougares F., Schwenk H., and Yoshua Bengio Y.. 2014. Learning phrase representations using RNN encoder–decoder forstatisticalmachinetranslation. In Proceedings of the 2014 Conference on Empirical Methods in Natural Language Processing. Doha, Qatar, pages1724– 1734.

\bibitem{ref_article19}
Luong T., Pham H. and Manning C.D. 2015. Effective approaches to attention-based neural machine translation. In Proceedings of the 2015 Conference on Empirical Methods in Natural Language Processing. Lisbon, Portugal, pages 1412–1421.

\bibitem{ref_article20}
Bahdanau D., Cho K. and Bengio Y. 2015. Neural machine translation by jointly learning to align and translate. In Proceedings of the 3rd International Conference on Learning Representations. San Diego, CA, USA.

\bibitem{ref_article21}
Kingma D.P. and Ba J. 2014. Adam: A method for stochastic optimization. CoRR, abs/1412.6980

\bibitem{ref_article22}
Devlin, J., Chang, M.W., Lee, K., and Toutanova, K. 2018. Bert: Pretraining of deep bidirectional transformers for language understanding. arXiv preprint arXiv:1810.04805 .

\end{thebibliography}
%

\end{document}